\documentclass[review]{elsarticle}

\usepackage{lineno,hyperref}
\usepackage{amsmath}
\usepackage{times}
\usepackage{epsfig}
\usepackage{graphicx}
\usepackage{amsmath}
\usepackage{amssymb}

\usepackage{caption}
\usepackage{algorithm}
\usepackage{algorithmic}
\usepackage{float}
\usepackage{amsmath}
\usepackage{mathrsfs}

\journal{Journal of \LaTeX\ Templates}

\begin{document}

\begin{frontmatter}

\title{Structure fusion based on graph convolutional networks for semi-supervised classification}

\author[mymainaddress]{Guangfeng Lin}\corref{mycorrespondingauthor}
\cortext[mycorrespondingauthor]{Corresponding author}
\ead{lgf78103@xaut.edu.cn}

\author[mymainaddress]{Jing Wang}
\author[mymainaddress]{Kaiyang Liao}
\author[mymainaddress]{Fan Zhao}
\author[mymainaddress]{Wanjun Chen}


\address[mymainaddress]{Information Science Department, Xi'an University of Technology,\\
 5 South Jinhua Road, Xi'an, Shaanxi Province 710048, PR China}


\begin{abstract}
Suffering from the multi-view data diversity and complexity for semi-supervised classification, most of existing graph convolutional networks focus on the networks architecture construction or the salient graph structure preservation, and ignore the the complete graph structure for semi-supervised classification contribution. To mine the more complete distribution structure from multi-view data with the consideration of the specificity and the commonality, we propose structure fusion based on graph convolutional networks (SF-GCN) for improving the performance of semi-supervised classification. SF-GCN can not only retain the special characteristic of each view data by spectral embedding, but also capture the common style of multi-view data by distance metric between multi-graph structures. Suppose the linear relationship between multi-graph structures, we can construct the optimization function of structure fusion model by balancing the specificity loss and the commonality loss. By solving this function, we can simultaneously obtain the fusion spectral embedding from the multi-view data and the fusion structure as adjacent matrix to input graph convolutional networks for semi-supervised classification. Experiments demonstrate that the performance of SF-GCN outperforms that of the state of the arts on three challenging datasets, which are Cora,Citeseer and Pubmed in citation networks.
\end{abstract}


\end{frontmatter}

\section{Introduction}
As a efficient representation of data distribution, graph plays a important role for describing the intrinsic structure of data. Therefore, many existing works have constructed the significant theory and method depending on the graph structure of data in pattern recognition, such as graph cut building energy function for semantic segmentation task \cite{veksler2019efficient}, graph-based learning system constructing the accurate recommendations for the interaction of the different objects \cite{monti2017geometric} \cite{ying2018graph}, graph modeling molecules bioactivity for drug discovery \cite{defferrard2016convolutional} \cite{gilmer2017neural} and graph simulating the link connection of citation network for the different group classification \cite{defferrard2016convolutional}\cite{gilmer2017neural}\cite{khan2019multi}. In fact, we usually observe objects and their relationship (\textbf{\textsl{this relationship is defined as the objects of structure, which often can be described by graph.}}) from multiple views, which provide the more abundant and complete information for object recognition. Learning on multi-graph (multiple observation structure) can effectively mine multiple relationship to discriminate the different object.

\begin{figure*}[ht]
  \begin{center}
\includegraphics[width=1\linewidth]{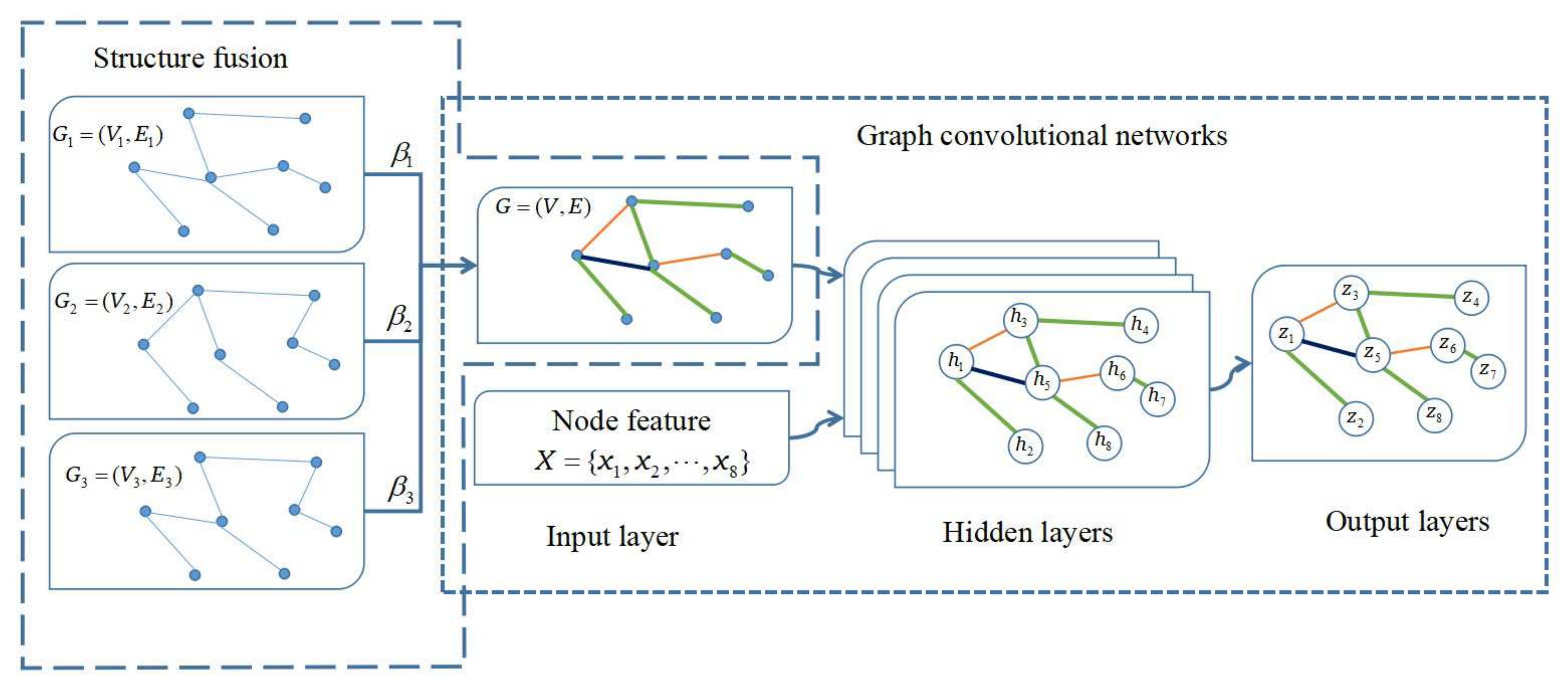}
\end{center}
\vspace{-0.2in}
 \caption{The diagram of structure fusion based on graph convolutional networks (SF-GCN), in which three graphs indicating the structure of the multi-view data and eight nodes (the different color connecting lines mean the various connecting weights) expressing the multi-node in these graphs;$\beta=[\beta_{1}\ \beta_{2}\ \beta_{3}]$ showing the linear coefficient between multi-graph structure for complementary fusion.}
  \label{SF-GCN1}
 \end{figure*}

Existing learning methods on multi-graph trend to tow ways. One is structure fusion \cite{Lin20131286} \cite{Lin2014146}\cite{7268821}\cite{7301305}\cite{Lin20161}\cite{Lin2017275}\cite{Lin2017Dynamic}\cite{Lin2018structure} \cite{lin2018class}\cite{LINGF2018}\cite{lin2019transfer} or diffusion on tensor product graph\cite{yang2011affinity} \cite{yang2012affinity} \cite{bai2019automatic} \cite{li2019semi} \cite{bai2017regularized} \cite{bai2017ensemble} based on the complete data, which include each view observation data. Another is graph convolutional networks for the salient graph structure preservation \cite{khan2019multi} based on the incomplete data, which lost some view observation data. For example, link relationship can be extracted by application necessary in citation networks, but it can not be described by the corresponding observation data computation. In other words, these link relationship exists, while the corresponding support data lost. Therefore, the method based on graph convolutional networks usually ignores the complete complementary of the different observation structure based on the incomplete multi-view data. To analysis this issue, we attempt to construct structure fusion based on graph convolutional networks for classification. Figure \ref{SF-GCN1} shows the overall flow diagram of structure fusion based on graph convolutional networks (SF-GCN). The inspiration of SF-GCN comes from Multi-GCN in the literature \cite{khan2019multi}, but there are tow points difference comparison with Multi-GCN. One is that SF-GCN considers the inequality of multiple structures, while Multi-GCN only equally deal with their relationship. The other is that SF-GCN focuses on the contributions of all nodes structure in the fusion structure, while Multi-GCN only emphasises on the salient structure of the part nodes. From the classification sense,the strong and weak links between nodes both considered for complementing structure can  more fit to the intrinsic structure of the data for classification.

Our contributions can be summarized as following. (a)We present a novel structure fusion based on graph convolutional networks (SF-GCN) that discriminates the different classes by optimizing the linear relationship of multiple observation structure with balancing the specificity loss and the commonality loss. (b) In three citation datasets with document sparse feature and document link relationship, the proposed SF-GCN outperforms the state of the arts for semi-supervised classification. (c) Our model is generalized the different multi-graph fusion methods for evaluating the performance of the proposed SF-GCN.

\section{Related Works}
In this section, we mainly review recent related works about structure fusion and graph neural network.

\subsection{Structure fusion}
Structure fusion initially proposed in \cite{Lin20131286}can merge multiple structures for shape classification. In the follow-up works, the extend methods can be divided into three categories according to the different fusion ways. The first kind of methods try to find the optimized linear relationship of multiple observation structure based on the different manifold learning method \cite{Lin2014146} or statistics model analysis\cite{7268821}. The second kind of methods attempt to mine the nonlinear relationship of heterogeneous feature structure based on the global feature\cite{7301305}\cite{Lin20161} or the local feature encoding \cite{Lin2017275}. The third kind of methods can capture the dynamic changes of multiple structures for semi-supervised classification \cite{Lin2017Dynamic}or the structure propagation way for zero-shot learning\cite{Lin2018structure} \cite{lin2018class} \cite{LINGF2018}\cite{lin2019transfer}.

From above mention, existing methods emphasis on the completeness of data and their relationship based on data project, while graph convolutional networks focus on the transformation and evolution of data structure by deep learning frameworks. Therefore, we expect to draw support from structure fusion based on structure metric and graph convolutional networks for processing the incomplete view data, and find evolution law of the the fusion structure with the consideration of their specificity and commonality.

\subsection{Graph neural network}
Graph neural networks can discover the potential data relationship by the computation based on graph nodes and links. Especially, the computation is defined as convolution for graph data, and graph convolution networks (GCN) have become a promising direction in pattern recognition. In terms of the different node representation, graph convolution networks include spectral-based GCN and spatial-based GCN. Spectral-based GCN can define graph Fourier transform based on graph Laplacian matrix for projecting graph signal into the orthonormal space. The difference of these methods is the selection of the filter, which may be the learned parameters set\cite{bruna2013spectral}, Chebyshev polynomial \cite{defferrard2016convolutional}, or the first-order Chebyshev polynomial\cite{kipf2016semi} \cite{chen2018fastgcn}. Spatial-based GCN regards images as a special graph with a pixel describing a node. To avoid the storage of all states, these methods have present the improved training strategies, such as sub-graph training \cite{hamilton2017inductive}or stochastically asynchronous training \cite{dai2018learning}. Furthermore, some complex networks architecture can utilize gating unite to control the selection of node neighborhood\cite{liu2018geniepath}, or design two graph convolution networks with the consideration of the local and global consistency on graph\cite{zhuang2018dual} , or adjust the receptive field of node on graph by hyper-parameters \cite{van2018filter}.

Because spectral-based GCN can explicitly construct the learning model on the graph structure,which can easily be separated from GCN architecture. Therefore, this point provides a way for processing multiple structures, which may be incremental. In this paper, we focus on the important role of graph (structure) from multi-view data, and attempt to mine the plentiful information from multiple structures for spectral-based GCN inputting.

\section{Structure fusion based on graph convolutional networks}

 \begin{figure*}[ht]
  \begin{center}
\includegraphics[width=1\linewidth]{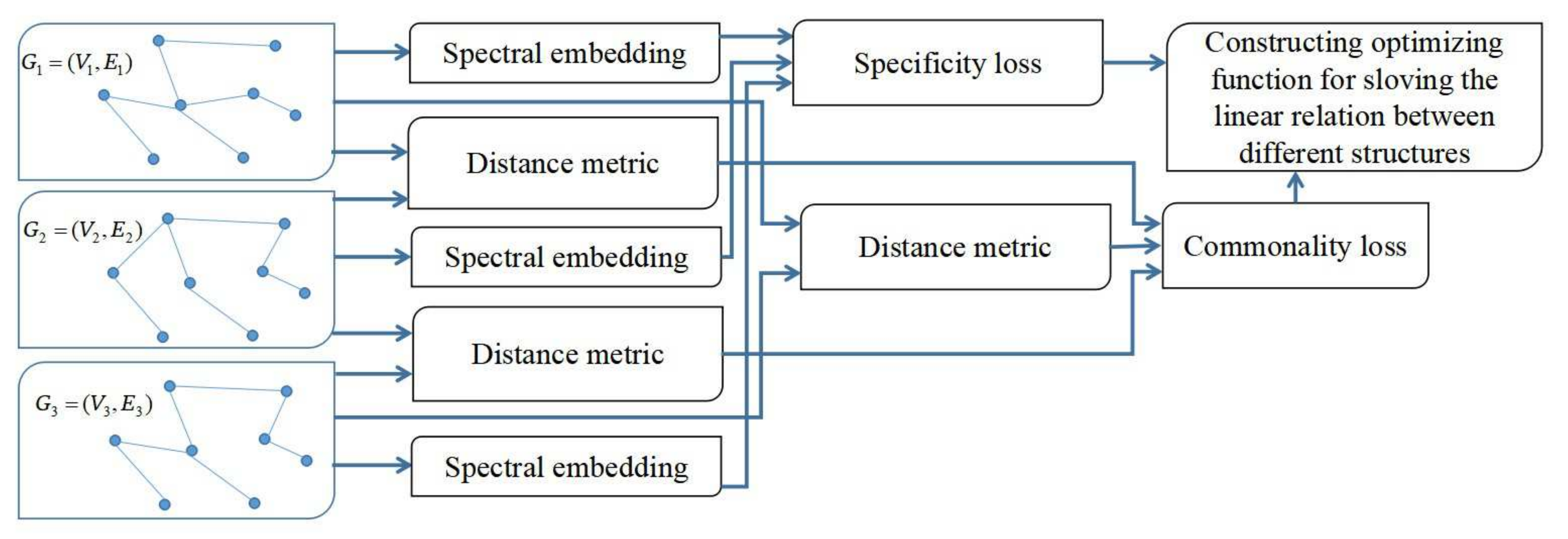}
\end{center}
\vspace{-0.2in}
 \caption{The mechanism of structure fusion in SF-GCN.}
  \label{SF-GCN2}
 \end{figure*}

To the best of our knowledge,existing structure fusion methods usually construct the optimizing function for feature projection, in which feature data and the corresponding structure jointly participate in computation. Because of the possible loss and the structure preservation of multi-view data, we expect to build a novel structure fusion by structure metric, in which the optimizing function only involves multiple structures for avoiding the negative effect of the data lost. Simultaneously, multiple structures have each specificity and their commonality. Therefore,we also anticipate that a novel structure fusion can be constrained by these characteristics of multiple structures. Figure \ref{SF-GCN2} demonstrates the internal mechanism of structure fusion in SF-GCN. First, we construct the specificity loss based on spectral embedding method with the consideration of multiple structure linear relationship. Second, we measure the commonality loss between multiple structures based on distance metric in Grassman manifold. Finally, we jointly exploit the structure fusion based on two losses, and input GCN for classification.

\subsection{Specificity loss of multiple structures}
Given an object set with $m$ multi-view, we can use graph $G_{i}$ to describe the observation distribution of data on each view. Therefore, the graph $G_{i}$ is the representation of the observation structure and $G=\{G_{i}|i=1,2,...,m\}$ can indicate multiple structures of data from multi-view observations. Because multiple structures detail the same object set, each $G_{i}$ includes the same vertex set $V$, or the possible different edges set $E_{i}$. If $W_{i}$ is the adjacency matrix of $G_{i}$ and is the numerical expression of the structure in $i$th view. In terms of spectral embedding, we can obtain the following optimization function on the embedding matrix $Y_{i}\in R^{n\times k}$ ($n$ is the number of samples, and $k$ is the dimension of embedding space) of each view.
\begin{align}
\label{loss1}
\begin{aligned}
&Y_{i}=\arg \min tr(Y_{i}^{T}L_{i}Y_{i}),
&s.t.\ Y_{i}^{T}Y_{i}=\mathbf{1}
 \end{aligned}
\end{align}

Where, $L_{i}=D_{i}-W_{i}$ is Laplacian matrix of $G_{i}$, $D_{i}$ is the degree matrix for $G_{i}$. Therefore, $L_{i}$ can still describe the characteristic of structure on graph $G_{i}$. We can compute the embedding matrix $Y_{i}$ by optimizing equation (\ref{loss1}), which is equivalent to a eigenvalue solution problem. When all eigenvalues are solved, eigenvectors corresponding to the smallest eigenvalues can build the embedding matrix $Y_{i}$, which can project the original nodes into the low dimensional spectral space \cite{xia2010multiview}. We can regard $tr(Y_{i}^{T}L_{i}Y_{i})$ as the specificity loss of structure on graph graph $G_{i}$, and then we can reformulate the specificity loss of multiple structures as follow.
\begin{align}
\label{loss2}
\begin{aligned}
&Loss_{s}=tr(Y^{T}LY)
\end{aligned}
\end{align}
Where, $Y$ is the embedding matrix of multiple structures in graph $G$ and closely approximates $Y_{i}$. Suppose fusion structure $W$ is the linear combination of $W_{i}$, then $L$ and $L_{i}$ have the same linear relationship $L=\sum_{i=1}^{m}\beta_{i}L_{i}$, in which $\beta_{i}$ is the linear coefficient to encode the importance of multiple structures.

\subsection{Commonality loss of multiple structures}
\label{commonality}
To measure the commonality loss of multiple structures, we need metric the distance between Laplacian matrix $L_{i}$ and $L$.  According to the solvation of the equation (\ref{loss1}), we can obtain the equation (\ref{loss3}) for describing the internal connection between embedding matrix $Y_{i}$ and the corresponding Laplacian matrix $L_{i}$.
\begin{align}
\label{loss3}
\begin{aligned}
&L_{i}=Y_{i}\lambda_{i} Y_{i}^{T}
\end{aligned}
\end{align}
Where, $\lambda_{i}$ is diagonal matrix, in which diagonal values is the smallest eigenvalues of $L_{i}$. $Y_{i}$ can be explained a subspace for preserving the smaller variance of the column in $L_{i}$, that is for reserving the bigger variance of the column in the structure $W_{i}$. In other words, $Y_{i}$ can keep the more discrimination of the data. Similarly, $Y$ has the same sense in multiple observation structure. Therefore, we can replace the distance between Laplacian matrix $L_{i}$ and $L$ by the distance between $Y_{i}$ and $Y$ for indirectly computing the commonality loss of multiple structures. This point is consistent with  the specificity loss of the assumption, which is that $Y_{i}$ approximates $Y$ between each view and multi-view.

In terms of Grassmann manifold theory \cite{lin2012multi}\cite{turaga2011statistical}, the orthonormal matrix $Y_{i}\in R^{n\times k}$ can be regard as the column of $Y_{i}$ spanning an unique subspace, which can be project into an unique point on Grassmann manifold $\mathcal{G}(n,k)$. Similarly, $Y$ also can be mapped into an unique point on this Grassmann manifold. Therefore, the principle angles $\{\theta_{j}\}_{j=1}^{k}$ between these subspaces can represent the distance between $Y_{i}$ and $Y$. Furthermore, this distance can be reformulate as following\cite{dong2013clustering}.
\begin{align}
\label{loss4}
\begin{aligned}
&d^{2}(Y,Y_{i})=\sum_{j=1}^{k}\sin^{2}\theta_{j}=k-tr(YY^{T}Y_{i}Y_{i}^{T})
\end{aligned}
\end{align}
In multiple structures, we can define  the commonality loss $Loss_{c}$ as the distance between $Y$ and $\{Y_{i}\}_{i=1}^{m}$ as following.
\begin{align}
\label{loss5}
\begin{aligned}
&Loss_{c}=\sum_{i=1}^{m}d^{2}(Y,Y_{i})=km-\sum_{i=1}^{m}tr(YY^{T}Y_{i}Y_{i}^{T})
\end{aligned}
\end{align}
\subsection{Structure fusion by structure metric losses}
\label{SF}
As two structure metric losses, specificity loss can balance the contribution of the structure in each view, while commonality loss can consider the similarity of multiple structures in multi-view. These structure metric losses can both constrain the linear relationship $\{\beta_{i}\}_{i=1}^{m}$ of multiple structures. Therefore, we combine these structure metric losses as a total loss for encoding the importance of multiple structures. The total loss can be reformulated as following.
\begin{align}
\label{loss6}
\begin{aligned}
&Loss=loss_{s}+\alpha loss_{c}
\end{aligned}
\end{align}
Where, $\alpha$ is regularization parameter.  From equation (\ref{loss6}), we can construct the object optimization function as following.
\begin{align}
\label{loss7}
\begin{aligned}
\{Y,\{\beta_{i}\}_{i=1}^{m}\}&=\arg \min (loss_{s}+\alpha loss_{c})\\
&=\arg \min tr(Y^{T}LY)+\alpha(km-\sum_{i=1}^{m}tr(YY^{T}Y_{i}Y_{i}^{T}))\\
&=\arg \min tr(Y^{T}\sum_{i=1}^{m}\beta_{i}L_{i}Y)+\alpha(km-\sum_{i=1}^{m}tr(YY^{T}Y_{i}Y_{i}^{T}))\\
&s.t.\ Y^{T}Y=\mathbf{1}, \ \sum_{i=1}^{m}\beta_{i}=1,\ \alpha >0
\end{aligned}
\end{align}
In commonality loss, constant term $km$ can not influence the loss trend change, so we may remove this term for conveniently computing. Equation (\ref{loss7}) is reformulated as equation (\ref{loss8}) for balancing parameter $\{\beta_{i}\}_{i=1}^{m}$ between $0$ and $1$.
\begin{align}
\label{loss8}
\begin{aligned}
\{Y,\gamma\}&=\arg \min (tr(Y^{T}\sum_{i=1}^{m}\beta_{i}L_{i}Y)-\alpha\sum_{i=1}^{m}tr(YY^{T}Y_{i}Y_{i}^{T})))^2\\
&=\arg \min (tr(Y^{T}(\sum_{i=1}^{m}\beta_{i}L_{i}-\alpha\sum_{i=1}^{m}Y_{i}Y_{i}^{T})Y))^2\\
&s.t.\ Y^{T}Y=\mathbf{1},\ \sum_{i=1}^{m}\beta_{i}=1,\ \alpha >0,\ \gamma=\{\{\beta_{i}\}_{i=1}^{m},\alpha\}
\end{aligned}
\end{align}
Equation (\ref{loss8}) is a nonconvex optimization problem, we can solve this problem by $Y$ and $\gamma$ alternated optimization. If $\gamma$ is fixed, equation (\ref{loss8}) can be transformed as a eigenvalue solving problem as following.
\begin{align}
\label{loss9}
\begin{aligned}
\{Y\}&=\arg \min tr(Y^{T}MY)\\
&s.t.\ Y^{T}Y=\mathbf{1},\ \alpha >0,\ M=(\sum_{i=1}^{m}\beta_{i}L_{i}-\alpha\sum_{i=1}^{m}Y_{i}Y_{i}^{T}),\ \gamma=\{\{\beta_{i}\}_{i=1}^{m},\alpha\}
\end{aligned}
\end{align}
Equation (\ref{loss9})is equivalent to a eigenvalue solution problem. When all eigenvalues of $M$ are solved, eigenvectors corresponding to the smallest eigenvalues can build the fusion embedding matrix $Y$. If $Y$ is fixed, equation (\ref{loss8}) can be converted into a quadratic programming problem as following.
\begin{align}
\label{loss10}
\begin{aligned}
\{\gamma\}&=\arg \min (tr(Y^{T}(\sum_{i=1}^{m}\beta_{i}L_{i}-\alpha\sum_{i=1}^{m}Y_{i}Y_{i}^{T})Y))^2\\
&s.t.\ \sum_{i=1}^{m}\beta_{i}=1,\ \alpha >0,\ \gamma=\{\{\beta_{i}\}_{i=1}^{m},\alpha\}
\end{aligned}
\end{align}
By alternated solving between equation (\ref{loss9}) and equation (\ref{loss10}), we can obtain fusion embedding matrix $Y$ and the linear relationship $\gamma$ of multiple structures. Furthermore, fusion structure (fusion adjacent matrix) can be computed by $W=\sum_{i=1}^{m}\beta_{i}W_{i}$.

Algorithm \ref{alg1} shows the pseudo code for fusion structure of multiple structures. In this algorithm, there are three steps. The first step (line 1) initializes the linear relationship of multiple structures. The second step(from line 2 to line 3) computes the Laplacian matrix and the spectral embedding in each view. The third step (from line 4 to line 6) alternately optimizes the spectral embedding fusion and the linear relationship of multiple structures. The last step (line 8) calculates fusion structure by the linear combination of each structure. Therefore, the complexity of this algorithm is $O(mn^{3}+mn^{2}kT+k^{3.5}l^{2}T)$, in which $m$ represents multi-view; $n$ is the sample number; $k$ is the dimension of the selected eigenvectors; $T$ is the iterative times of optimization; $l$ is the number of bits in the input of algorithm.

\begin{algorithm}[ht]
  \caption{Fusion structure of multiple structures}
 \begin{algorithmic}[1]
 \label{alg1}
\renewcommand{\algorithmicrequire}{\textbf{Input:}}
\renewcommand{\algorithmicensure}{\textbf{Output:}}
\renewcommand{\algorithmicreturn}{\textbf{Iteration:}}
   \REQUIRE $\{W_{i}\}_{i=1}^{m}$ : $n\times n$ adjacency matrices of graph $\{G_{i}\}_{i=1}^{m}$; $\alpha$: regularization parameter of the total loss; $T$: the iteration times
   \ENSURE $W$: fusion structure of multiple structures
   \STATE Initializing the linear relationship $\{\beta_{i}\}_{i=1}^{m}$ of multiple structures and regularization parameter $\alpha$
   \STATE Computing Laplacian matrix $L_{i}$ of $G_{i}$
   \STATE Computing the spectral embedding $Y_{i}$ of structure in each view by equation (\ref{loss1})
   \FOR {$1<t<T$}
   \STATE Computing the spectral embedding fusion $Y$ of multiple structures  in multi-view by equation (\ref{loss9})
   \STATE Updating the linear relationship $\gamma$ of multiple structures by equation (\ref{loss10})
   \ENDFOR
   \STATE Computing fusion structure by $W=\sum_{i=1}^{m}\beta_{i}W_{i}$
  \end{algorithmic}
\end{algorithm}

\subsection{Graph convolutional networks}
In terms of the multiplication of convolution in the Fourier domain, graph convolution is defined as the the multiplication between the signal $s\in R_{n}$ and the filter $g_{\eta}$\cite{bruna2013spectral}. Furtherly, graph convolution can also be approximated by $1^{th}$-order Chebyshev polynomials \cite{kipf2016semi} as following.

\begin{align}
\label{gcn1}
\begin{aligned}
g_{\eta}\ast s&=Ug_{\eta}U^{T}s\\
&\approx \sum_{k=0}^{1}\eta^{'}_{k} T_{k}(\tilde{L})s\\
&\approx \eta(I+\tilde{D}^{-1/2}\tilde{W}\tilde{D}^{-1/2})s
\end{aligned}
\end{align}
Where, $U$ is the eigen-decomposition of the normalized Laplacian $L=I-D^{-1/2}WD^{-1/2}$($I$ is the identity matrix; $D$ is the degree matrix of graph $G$); $\tilde{L}=I-\tilde{D}^{-1/2}\tilde{W}\tilde{D}^{-1/2}$ ($\tilde{D}$ and $\tilde{W}$ respectively are the rescaled degree and adjacent matrix  by $\tilde{W}=W+I$); $T_{k}$ expresses the Chebyshev polynomials; $\eta=\eta_{0}^{'}=-\eta_{1}^{'}$.

Fusion structure $W$ can directly be input into the above graph convolutional networks. The forward propagation based on two layers of graph convolutional networks can be indicated as following.
\begin{align}
\label{gcn2}
\begin{aligned}
Z=softmax(\tilde{W}ReLU(\tilde{W}S\Theta^{0})\Theta^{1})
\end{aligned}
\end{align}
Where, $Z$ is the output of networks;$S$ is the representation matrix of each nodes;$\Theta^{0}$ and $\Theta^{1}$ respectively are the $1^{th}$ and $2^{th}$ layer filter parameters; $ReLU$ and $softmax$ are the different type of activation function located in the various layers.

\section{Experiments}
For evaluating the proposed SF-GCN, we carry out the experiments from four aspects. Firstly, we conduct the comparing experiment between the proposed SF-GCN and the baseline methods, which include graph convolutional networks (GCN)\cite{kipf2016semi} with the combination view and Multi-GCN\cite{khan2019multi}. Secondly,we utilize the different multi-graph fusion methods for analyzing the intrinsic mechanism of the proposed SF-GCN. Thirdly, we show the experimental results between the proposed SF-GCN and the state of the art methods for the node classification in citation networks. Finally, we implement the proposed SF-GCN method of the lost structure for demonstrating the importance of the complete structure.
\subsection{Datasets}
We use the paper-citation networks of the citation networks in experiments. The three popular datasets usually utilized in node classification respectively are Cora, Citeseer and Pubmed. Cora dataset has $7$ classes that involve $2708$ the grouped publication about machine learning and their undirected graph. Citeseer dataset includes $6$ classes that have $3327$ scientific papers and their undirected graph. In these datasets, each publication stands for a node of the related graph and is represented by one-hot vector, each element of which can indicate the presence and absence state of a word in the learned directory. Pubmed dataset has $3$ classes that contain $19717$ diabetes-related publications and their undirected graph. In this dataset, each paper (each node of the related graph) can be described by a term frequency-inverse document frequency (TF-IDF)\cite{wu2019comprehensive}. Table \ref{table1} shows the statistics of these datasets. To obtain the structure of the second view from publication description, we normalize the cosine similarity between these publication. If these similarity is greater than $0.8$, we produce an edge for the corresponding to nodes in the citation network. This configuration is same in the literature \cite{khan2019multi}.

\begin{table*}[!ht]
\small
\renewcommand{\arraystretch}{1.0}
\caption{Three datasets statistics in citation networks. }
\label{table1}
\begin{center}
\newcommand{\tabincell}[2]{\begin{tabular}{@{}#1@{}}#2\end{tabular}}
\begin{tabular}{lp{1.5cm}p{1.5cm}p{1.5cm}p{1.5cm}p{1.5cm}}
\hline
\bfseries Datasets & \bfseries \tabincell{l}{Nodes \\number} & \bfseries \tabincell{l}{Edges \\number} & \bfseries \tabincell{l}{Classes \\number} & \bfseries \tabincell{l}{Feature \\dimension} &\bfseries \tabincell{l}{Label \\rate} \\
\hline \hline
Cora  & $2708$ &$5429$ &$7$ & $1433$ & $0.052$\\
\hline
Citeseer  & $3327$ &$4732$ &$6$ &$3703$ & $0.036$\\
\hline
Pubmed  & $19717$ &$44338$& $3$ &$500$& $0.003$\\
\hline
\end{tabular}
\end{center}
\end{table*}

\subsection{Experimental configuration}
In experiments, we follow the configuration in GCN\cite{kipf2016semi}, in which we train a two-layer GCN for maximum of $200$ epochs and test model in $1000$ labeled samples. Moreover, we select the same validation set of $500$ labeled sample for hyper-parameter optimization (dropout rate for all layers, number of hidden units and learning rate).

In proposed SF-GCN, we initially set the linear relationship $\{\beta_{i}\}_{i=1}^{m}$ of multiple structures and regularization parameter $\alpha$ as $0.5$, and then update these parameters in iteration optimization. The iteration time $T$ of the algorithm is $5$ according it's the convergence degree in fact.

\subsection{Comparison with the base-line methods}
\label{baseline}
The proposed method (SF-GCN) can be constructed based on GCN\cite{kipf2016semi}, and attempt to mine the different structure information for completing the intrinsic structure in multi-view data. Therefore, two base-line methods (GCN and Multi-GCN can find and capture the different structure information from the different consideration.) is involved for processing multi-view data based on GCN. GCN for multi-view\cite{kipf2016semi} can concatenate the different structure to build a sparse block-diagonal matrix where each block corresponding to the different structure (the adjacent matrix of different graph). Multi-GCN \cite{khan2019multi} can preserve the significant structure of the different structure by manifold ranking. In contrast with these base-line methods, the proposed method (SF-GCN) can not only enhance the common structure, but also retain the specific structure by structure fusion.

Table \ref{table2} shows that the classification performance of SF-GCN outperforms that of the base-line methods and the least improvement of SF-GCN respectively is $0.7\%$ for Cora, $2.1\%$ for Citeseer and $0.6\%$ for PubMed. However, GCN for multi-view is not superior to GCN for single-view, and it demonstrates that information mining of multi-view data is a key point for node classification. Therefore, SF-GCN attempt to mine the structure information from multi-view data for this purpose and obtain the better performance.

\begin{table}[!ht]
\small
\renewcommand{\arraystretch}{1.0}
\caption{Accuracy comparison of SF-GCN method with the base-line methods for node classification in citation network. View1 stands for graph structure from the original dataset, while view2 indicates graph structure from the cosine similarity of node representation}
\label{table2}
\begin{center}
\newcommand{\tabincell}[2]{\begin{tabular}{@{}#1@{}}#2\end{tabular}}
\begin{tabular}{lp{1cm}p{1cm}p{1cm}p{1cm}}
\hline
\bfseries Method &\bfseries Cora &\bfseries Citeseer &\bfseries PubMed  \\
\hline \hline
GCN \cite{kipf2016semi} for view1  & $81.5$   &$70.3$ & $78.7$    \\
\hline
GCN \cite{kipf2016semi} for veiw2  & $53.6$   &$50.7$ & $69.5$   \\
\hline
GCN \cite{kipf2016semi} for multi-view  & $80.4$   &$70.7$ & $78.2$  \\
\hline
Multi-GCN  \cite{khan2019multi} & $82.5$   &$71.3$  &NA  \\
\hline\hline
SF-GCN  & $\textbf{83.3}$   &$\textbf{73.4}$  & $\textbf{79.3}$  \\
\hline
\end{tabular}
\end{center}
\end{table}

\subsection{structure fusion generalization}
\label{generalization}
Structure fusion (SF) focuses on the complementation of the distribution structure from the different view data, and $W=\sum_{i=1}^{m}\beta_{i}W_{i}$ can be defined in section \ref{SF}.  However, the diffusion \cite{yang2012affinity} \cite{bai2017regularized} \cite{bai2017regularized} and propagation\cite{LINGF2018} \cite{Lin2018structure} of the different structure can also describe the complex relationship of the various structure, and become the important part of structure fusion. Therefore, we can define fusion structure $W$ by the propagation fusion (PF) of the different structure as follow.
\begin{align}
\label{pf1}
\begin{aligned}
W=\prod_{i}^{m}W_{i}
\end{aligned}
\end{align}
The propagation fusion can exchange and interact the relationship information between the various structures, and mine the neighbour relationship of multiple structures. However,this propagation can effect on the clustering performance of the original structure by high-order iteration multiplication. Therefore, we only consider zero-order (for example SF) and first-order (for instance PF) multiplication, that is structure propagation fusion (SPF) as follow.
\begin{align}
\label{pf2}
\begin{aligned}
W=\sum_{i=1}^{m}\beta_{i}W_{i}+\prod_{i}^{m}W_{i}
\end{aligned}
\end{align}
For evaluating structure fusion generalization, we compare structure fusion based graph convolutional networks (SF-GCN), propagation fusion based graph convolutional networks (PF-GCN) and structure propagation fusion based graph convolutional networks (SPF-GCN).In Table \ref{table3}, we observe that the performance of SPF-GCN is better than that of other method, and the least improvement of SPF-GCN respectively is  $0.2\%$ for Cora, $0.1\%$ for Citeseer and $0.7\%$ for PubMed, while the performance of SP is superior to that of PF-GCN, and the improvement of SF-GCN respectively is  $0.6\%$ for Cora, $0.9\%$ for Citeseer and $0.2\%$ for PubMed Therefore, PF and SF both are benefit for further mining the structure information and the role of SF is more important than that of PF.

\begin{table}[!ht]
\small
\renewcommand{\arraystretch}{1.0}
\caption{Structure fusion generalization classification accuracy in three methods, which are structure fusion based graph convolutional networks (SF-GCN), propagation fusion based graph convolutional networks (PF-GCN) and structure propagation fusion based graph convolutional networks (SPF-GCN)}
\label{table3}
\begin{center}
\newcommand{\tabincell}[2]{\begin{tabular}{@{}#1@{}}#2\end{tabular}}
\begin{tabular}{lp{1cm}p{1cm}p{1cm}p{1cm}}
\hline
\bfseries Method &\bfseries Cora &\bfseries Citeseer &\bfseries PubMed  \\
\hline \hline
SF-GCN & $83.3$   &$73.4$ & $79.3$    \\
\hline
PF-GCN  & $82.7$   &$72.5$ & $79.1$   \\
\hline\hline
SPF-GCN  & $\textbf{83.5}$   &$\textbf{73.5}$  & $\textbf{80.0}$  \\
\hline
\end{tabular}
\end{center}
\end{table}

\subsection{Comparison with the state-of-the-arts}
\label{State-of-the-arts}
Because graph convolutional networks and structure fusion are basic ideas for constructing the proposed method SPF-GCN, we analyze six related state-of-the-arts methods for evaluating SPF-GCN. These methods include two categories. One is node neighbour information exploiting for GCN, and another is node information fusion based on GCN.

Node neighbour information exploiting attempts to capture the distribution structure of the node neighbour for obtain the stable graph structure representation. For example, graph attention networks(GAT) can specify different weights to different nodes in a neighborhood \cite{Veli2017Graph}; stochastic training of graph convolutional networks (StoGCN) allows sampling an arbitrarily small neighbor size \cite{2017arXiv171010568C}; deep graph infomax(DGI) can maximize mutual information between different level subgraph centered around nodes of interest (the different way for considering neighbour information)\cite{Veli2018Deep}.

Node information fusion tries to mine the information from multi-view node description or multiple structures for complementing the difference of multi-view data. For instance, large-scale learnable graph convolutional networks (LGCN) can fuse neighbouring nodes feature by ranking selection to transform graph data into grid-like structures in 1-D format\cite{2018arXiv180803965G}; dual graph convolutional networks (DGCN) can consider local and global consistency for fusion different views graph of raw data\cite{zhuang2018dual}; Multi-GCN can extract and select the significant structure form multi-view structure by manifold ranking\cite{khan2019multi}.

The proposed method SPF-GCN belongs to node information fusion method, and the difference compared with the above methods focuses on the complementary of multiple structures by mining their commonality, specificity and interactive propagation. Table \ref{table4} shows SPF-GCN outperforms other state-of-the-art methods except DGCN in Cora and PubMed datasets. Although SPF-GCN and DGCN reach to the same performance in Cora and PubMed datasets, SPF-GCN can preserve the higher computation efficient of the original GCN because of the separable computation between structure fusion and GCN.

\begin{table}[!ht]
\small
\renewcommand{\arraystretch}{1.0}
\caption{Accuracy comparison of SF-GCN and SPF-GCN with state-of-the-art methods for node classification in citation network.}
\label{table4}
\begin{center}
\newcommand{\tabincell}[2]{\begin{tabular}{@{}#1@{}}#2\end{tabular}}
\begin{tabular}{lp{2cm}p{2cm}p{2cm}p{2cm}}
\hline
\bfseries Method &\bfseries Cora &\bfseries Citeseer &\bfseries PubMed  \\
\hline \hline
GAT\cite{Veli2017Graph} & $83.0\pm0.7$   &$72.5\pm0.7$ & $79.3\pm0.3$    \\
\hline
StoGCN\cite{2017arXiv171010568C}  & $82.0\pm0.8$   &$70.9\pm0.2$ & $78.7\pm0.3$   \\
\hline
DGI\cite{Veli2018Deep}  & $82.3\pm0.6$   &$71.8\pm0.7$ & $76.8\pm0.6$   \\
\hline
LGCN\cite{2018arXiv180803965G}   & $83.3\pm0.5$   &$73.0\pm0.6$ & $79.5\pm0.2$   \\
\hline
DGCN\cite{zhuang2018dual}  & $\textbf{83.5}$   &$72.6$ & $\textbf{80.0}$   \\
\hline
Multi-GCN\cite{khan2019multi}  & $82.5$   &$71.3$ & NA   \\
\hline\hline
SF-GCN & $83.3$   &$73.4$ & $79.3$   \\
\hline
SPF-GCN  & $\textbf{83.5}$   &$\textbf{73.5}$  & $\textbf{80.0}$  \\
\hline
\end{tabular}
\end{center}
\end{table}

\subsection{Incomplete structure influence}
Structure fusion can capture the complementary information of multiple structures, and this complementary information can supply an efficient way for incomplete structure influence. The main reason of the incomplete structure may be because of noise and data loss in practical situation. For evaluating the performance of the proposed methods under the condition of the incomplete structure, we design a experiment in all datasets. In semi-supervised classification, the distribution structure of test datasets is more important than that of train datasets, and can assure the performance of classification because of the transfer relation of structure between train and test datasets. Therefore, we delete the some structure of test datasets to destroy this transfer relation for simulating incomplete structure.

\begin{figure*}[ht]
 \begin{center}
\includegraphics[width=1\linewidth]{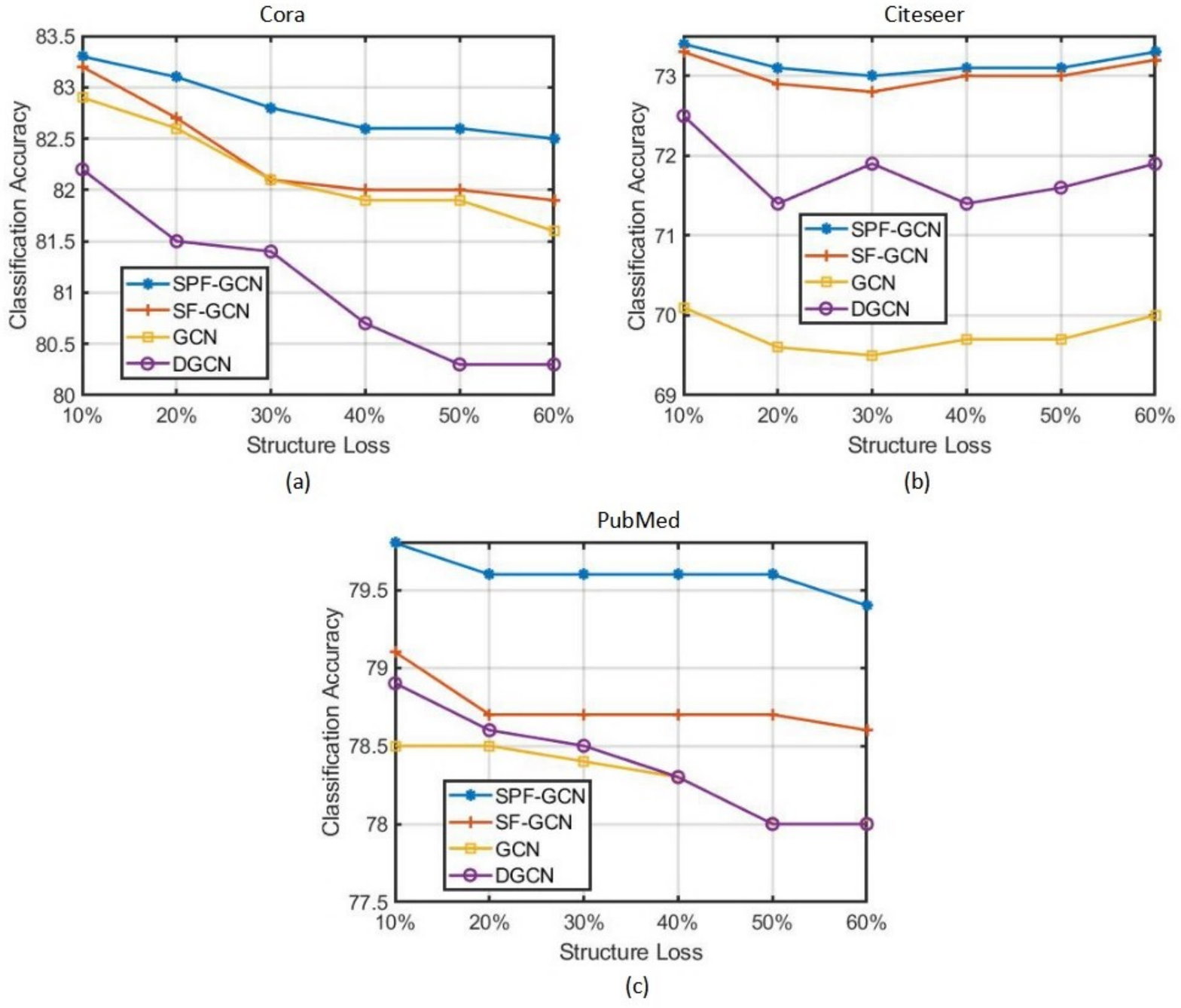}
\vspace{-0.2in}
 \caption{Impact of structure loss on classification accuracy for citation networks on (a) Cora,(b)Citeseer and (c)PubMed datasets.}
  \label{fig3}
  \end{center}
 \end{figure*}

In the details, we proportionally set the adjacency matrix(graph structure from the original dataset) of elements (corresponding to test datasets) to zero from $10\%$ to $60\%$, and then respectively implement GCN for multi-view \cite{kipf2016semi}, DGCN \cite{zhuang2018dual}, SF-GCN and SPF-GACN methods in all dataset. In figure \ref{fig3}, we select structure loss degree from $10\%$,$20\%$,$30\%$,$40\%$,$50\%$,$60\%$ to construct the different classification model for evaluating the performance of the compared methods. Especially, there is the smaller descent of SPF-GCN classification accuracy with structure loss increasing from $10\%$ to $60\%$, e.g. $83.3$ to $82.5$ on Cora, $73.4$ to $73.0$ on Citeseer and $79.8$ to $79.4$ on PubMed. We can observe that the proposed SF-GCN and SPF-GACN is more stable and robust with incomplete degree increasing of structure than GCN for multi-view and DGCN. In this situation, the performance of SPF-GACN is better than that of SF-GACN, while the performance of GCN outperforms that of DGCN in Cora datasets, and the performance of DGCN is superior to that of GCN in Citeseer and PubMed datasets. The details of this reason can be analyzed in section \ref{analysis}.

\subsection{Experimental results analysis}
\label{analysis}
In our experiments, we compare the proposed method with eight methods, which contain two kinds of base-line methods (Multi-GCN\cite{khan2019multi}, GCN \cite{kipf2016semi} for multi-view, GCN \cite{kipf2016semi} for view1 and view2 in section \ref{baseline}), two kinds of structure fusion generalization methods (PF-GCN and SF-GCN in section \ref{generalization}), and six kinds of the state-of-the-art methods(GAT\cite{Veli2017Graph}, StoGCN\cite{2017arXiv171010568C}, DGI\cite{Veli2018Deep}, LGCN\cite{2018arXiv180803965G}, DGCN\cite{zhuang2018dual} and Multi-GCN\cite{khan2019multi} in section \ref{State-of-the-arts}). These methods can utilize the graph structure mining based graph convolutional networks for semi-supervised classification by the different ways. In contrast to other methods, the proposed SF-GCN and SPF-GCN methods focus on the complementary relationship of multiple structures by the consideration of their commonality and specificity. Moreover, the proposed SPF-GCN method not only capture the optimization distribution of fusion structure, but also emphasize on the interactive propagation between the different structures. From the above experiments, we can observe several points as following.

\begin{itemize}
\item  The performance of SF-GCN is superior to that of the base-line methods (Multi-GCN\cite{khan2019multi}, GCN \cite{kipf2016semi} for multi-view, GCN \cite{kipf2016semi} for view1 and view2 in section \ref{baseline}). GCN \cite{kipf2016semi} constructs a general graph convolutional architecture by the first-order approximation of spectral graph convolutions for greatly improving the computation efficiency of graph convolutional networks, and also provides a feasible deep mining frameworks for effective semi-supervised classification. For using multiple structures, GCN for multi-view can input a sparse block-diagonal matrix, each block of which corresponding to the different structure. Therefore, the relationship of each block (the different structure) is ignored for GCN, and this point leads to the poor performance (in some times, the performance of GCN for multi-view is worse than that of GCN for view1) of GCN for multi-view. In contrast, Multi-GCN\cite{khan2019multi} can capture the relationship of the different structure to preserve the significant structure of merging subspace. However, Multi-GCN\cite{khan2019multi} neglects the optimizing fusion relationship of the different structure, while the proposed SF-GCN focuses on finding these relationship by jointly considering the commonality and specificity loss of multiple structure for obtaining the better performance of semi-supervised classification.
\item SPF-GCN shows the best performance in structure fusion generalization experiments, whereas the performance of SF-GCN is better than that of SP-GCN. The main reason is that SF-GCN emphasises on the complement information by the optimizing fusion relationship of the different structure, while SP-GCN trends to the interactive propagation by the diffusion influence between the different structures. The complement fusion play the more important role than the interactive propagation because of the specificity structure of individual view data, but both fusion and propagation can contribute the multiple structures mining for enhancing the the performance of semi-supervised classification.
\item The performance improvement of SPF-GCN compared with six kinds of the state-of-the-art methods is respectively different. The similar performance of SPF-GCN is shown in the comparison with LGCN and DGCN in Cora, DGCN in PubMed. Except these situation, the better improvement of SPF-GCN can be demonstrated in other methods. The main reason is that LGCN can emphasis on  neighboring nodes feature fusion for the stable node representation and DGCN can correlate the local and global consistency for complementing the different structures. The proposed SPF-GCN expects not only to capture the structure commonality for complementing the different information, but also to preserve the structure specificity for mining the discriminative information. Therefore, the proposed SPF-GCN can improve the classification performance in the most experiments. In the least, the proposed SPF-GCN have the similar performance than the best performance of other method in all experiments. In addition, the proposed SPF-GCN is based on GCN frameworks, so it has the efficient implementation like GCN. In experiments, the computation efficiency of the proposed SPF-GCN is the highest than that of the state-of-the-art methods (the details of the computation efficiency in section \ref{commonality}).
\item Structure shows the distribution of data, and is very important for learning GCN model. Incomplete structure can evaluate the robustness of the related GCN model. We select the classical GCN, the state-of-the-art DGCN, SF-GCN and SPF-GCN for the robust test. The proposed SPF-GCN shows the best performance in three datasets. In Cora, the performance of GCN is better than DGCN,while the performance of GCN is worse than DGCN in Citeseer and PubMed. It shows that local and global consistency for fusing graph information in DGCN trend to the unstable characteristic because of the tight constraint of incomplete structure consistency. The loose constraint of GCN for incomplete structure correlation leads to the worse performance. The proposed SPF-GCN can compromise these constrains for balancing the incomplete structure information by optimizing the weight of multiple structures, and also connect the different structure for complementing the different information. Therefore, the proposed SPF-GCN obtains the best performance in experiments.
\item The proposed SPF-GCN expect to mine the commonality and the specificity of multiple structures. The commonality describes the similarity characteristic of structures by Grassmann manifold metric, while the specificity narrates the difference characteristic of structures by spectral embedding. In the proposed method, the specificity is constructed based on the commonality. Therefore, we only execute the ablation experiment for preserving the commonality loss by deleting the specificity loss from the total loss. This experiment obtain the following performance, that is $82.6\%$ in Cora, $71.5\%$ in Citeseer and $78.9\%$ in PubMed. These results obviously are worse than the performance of the proposed SF-GCN and SPF-GCN, which can balance the commonality and specificity for mining the suited weight of multiple structures.

\end{itemize}

\section{Conclusion}
We have proposed structure fusion based on graph convolutional networks (SF-GCN) to address the multi-view data diversity and complexity for semi-supervised classification. SF-GCN can not only adapt spectral embedding to preserve the specificity of structure, but also model the relationship of the different structure to find the commonality of multiple structures by manifold metric. Furthermore, the proposed structure propagation fusion based graph convolutional networks(SPF-GCN) can combine structure fusion framework with structure propagation to generating the completer structure graph for improving the performance of semi-supervised classification. At last, the optimization learning of the SF-GCN can obtain both the suitable weight for the different structure and the merge embedding space. For evaluating the proposed  SF-GCN and SPF-GCN, we carry out the comparison experiments about the baseline methods, the different multi-graph fusion methods, the state of the art methods and the the lost structure analysis on Cora,Citeseer and Pubmed datasets. Experiment results demonstrate SF-GCN and SPF-GCN get the promising results in semi-supervised classification.

\section{Acknowledgements}
The authors would like to thank the anonymous reviewers for their insightful comments that help improve the quality of this paper. This work was supported by NSFC (Program No.61771386,Program No.61671376 and Program No.61671374), Natural Science Basic Research Plan in Shaanxi Province of China (Program No.2017JZ020).

\section*{References}

\bibliography{mybibfile}

\begin{thebibliography}{10}
\expandafter\ifx\csname url\endcsname\relax
  \def\url#1{\texttt{#1}}\fi
\expandafter\ifx\csname urlprefix\endcsname\relax\def\urlprefix{URL }\fi
\expandafter\ifx\csname href\endcsname\relax
  \def\href#1#2{#2} \def\path#1{#1}\fi

\bibitem{veksler2019efficient}
O.~Veksler, Efficient graph cut optimization for full crfs with quantized
  edges, IEEE transactions on pattern analysis and machine intelligence\href
  {http://dx.doi.org/10.1109/TPAMI.2019.2906204}
  {\path{doi:10.1109/TPAMI.2019.2906204}}.

\bibitem{monti2017geometric}
F.~Monti, M.~Bronstein, X.~Bresson, Geometric matrix completion with recurrent
  multi-graph neural networks, in: Advances in Neural Information Processing
  Systems, 2017, pp. 3697--3707.

\bibitem{ying2018graph}
R.~Ying, R.~He, K.~Chen, P.~Eksombatchai, W.~L. Hamilton, J.~Leskovec, Graph
  convolutional neural networks for web-scale recommender systems, in:
  Proceedings of the 24th ACM SIGKDD International Conference on Knowledge
  Discovery and Data Mining, ACM, 2018, pp. 974--983.

\bibitem{defferrard2016convolutional}
M.~Defferrard, X.~Bresson, P.~Vandergheynst, Convolutional neural networks on
  graphs with fast localized spectral filtering, in: Advances in neural
  information processing systems, 2016, pp. 3844--3852.

\bibitem{gilmer2017neural}
J.~Gilmer, S.~S. Schoenholz, P.~F. Riley, O.~Vinyals, G.~E. Dahl, Neural
  message passing for quantum chemistry, in: Proceedings of the 34th
  International Conference on Machine Learning-Volume 70, JMLR. org, 2017, pp.
  1263--1272.

\bibitem{khan2019multi}
M.~R. Khan, J.~E. Blumenstock, Multi-gcn: Graph convolutional networks for
  multi-view networks, with applications to global poverty, arXiv preprint
  arXiv:1901.11213.

\bibitem{Lin20131286}
G.~Lin, H.~Zhu, X.~Kang, C.~Fan, E.~Zhang, Multi-feature structure fusion of
  contours for unsupervised shape classification, Pattern Recognition Letters
  34~(11) (2013) 1286 -- 1290.

\bibitem{Lin2014146}
G.~Lin, H.~Zhu, X.~Kang, C.~Fan, E.~Zhang, Feature structure fusion and its
  application, Information Fusion 20 (2014) 146 -- 154.

\bibitem{7268821}
G.~Lin, H.~Zhu, X.~Kang, Y.~Miu, E.~Zhang, Feature structure fusion modelling
  for classification, IET Image Processing 9~(10) (2015) 883--888.

\bibitem{7301305}
G.~Lin, G.~Fan, L.~Yu, X.~Kang, E.~Zhang, Heterogeneous structure fusion for
  target recognition in infrared imagery, in: 2015 IEEE Conference on Computer
  Vision and Pattern Recognition Workshops (CVPRW), 2015, pp. 118--125.

\bibitem{Lin20161}
G.~Lin, G.~Fan, X.~Kang, E.~Zhang, L.~Yu, Heterogeneous feature structure
  fusion for classification, Pattern Recognition 53 (2016) 1 -- 11.

\bibitem{Lin2017275}
G.~Lin, C.~Fan, H.~Zhu, Y.~Miu, X.~Kang, Visual feature coding based on
  heterogeneous structure fusion for image classification, Information Fusion
  36 (2017) 275 -- 283.

\bibitem{Lin2017Dynamic}
G.~Lin, K.~Liao, B.~Sun, Y.~Chen, F.~Zhao, Dynamic graph fusion label
  propagation for semi-supervised multi-modality classification, Pattern
  Recognition 68 (2017) 14--23.

\bibitem{Lin2018structure}
G.~Lin, Y.~Chen, F.~Zhao, Structure propagation for zero-shot learning, arXiv
  preprint arXiv:1711.09513.

\bibitem{lin2018class}
G.~Lin, C.~Fan, W.~Chen, Y.~Chen, F.~Zhao, Class label autoencoder for
  zero-shot learning, arXiv preprint arXiv:1801.08301.

\bibitem{LINGF2018}
G.~Lin, Y.~Chen, F.~Zhao, Structure fusion and propagation for zero-shot
  learning, in: Chinese Conference on Pattern Recognition and Computer Vision
  (PRCV), Springer, 2018, pp. 465--477.

\bibitem{lin2019transfer}
G.~Lin, W.~Chen, K.~Liao, X.~Kang, C.~Fan, Transfer feature generating networks
  with semantic classes structure for zero-shot learning, arXiv preprint
  arXiv:1903.02204.

\bibitem{yang2011affinity}
X.~Yang, L.~J. Latecki, Affinity learning on a tensor product graph with
  applications to shape and image retrieval, in: CVPR 2011, IEEE, 2011, pp.
  2369--2376.

\bibitem{yang2012affinity}
X.~Yang, L.~Prasad, L.~J. Latecki, Affinity learning with diffusion on tensor
  product graph, IEEE transactions on pattern analysis and machine intelligence
  35~(1) (2012) 28--38.

\bibitem{bai2019automatic}
S.~Bai, Z.~Zhou, J.~Wang, X.~Bai, L.~J. Latecki, Q.~Tian, Automatic ensemble
  diffusion for 3d shape and image retrieval, IEEE Transactions on Image
  Processing 28~(1) (2019) 88--101.

\bibitem{li2019semi}
Q.~Li, S.~An, L.~Li, W.~Liu, Semi-supervised learning on graph with an
  alternating diffusion process, arXiv preprint arXiv:1902.06105.

\bibitem{bai2017regularized}
S.~Bai, X.~Bai, Q.~Tian, L.~J. Latecki, Regularized diffusion process for
  visual retrieval, in: Proceedings of the Thirty-First AAAI Conference on
  Artificial Intelligence, AAAI Press, 2017, pp. 3967--3973.

\bibitem{bai2017ensemble}
S.~Bai, Z.~Zhou, J.~Wang, X.~Bai, L.~Jan~Latecki, Q.~Tian, Ensemble diffusion
  for retrieval, in: Proceedings of the IEEE International Conference on
  Computer Vision, 2017, pp. 774--783.

\bibitem{bruna2013spectral}
J.~Bruna, W.~Zaremba, A.~Szlam, Y.~LeCun, Spectral networks and locally
  connected networks on graphs, arXiv preprint arXiv:1312.6203.

\bibitem{kipf2016semi}
T.~N. Kipf, M.~Welling, Semi-supervised classification with graph convolutional
  networks, arXiv preprint arXiv:1609.02907.

\bibitem{chen2018fastgcn}
J.~Chen, T.~Ma, C.~Xiao, Fastgcn: fast learning with graph convolutional
  networks via importance sampling, arXiv preprint arXiv:1801.10247.

\bibitem{hamilton2017inductive}
W.~Hamilton, Z.~Ying, J.~Leskovec, Inductive representation learning on large
  graphs, in: Advances in Neural Information Processing Systems, 2017, pp.
  1024--1034.

\bibitem{dai2018learning}
H.~Dai, Z.~Kozareva, B.~Dai, A.~Smola, L.~Song, Learning steady-states of
  iterative algorithms over graphs, in: International Conference on Machine
  Learning, 2018, pp. 1114--1122.

\bibitem{liu2018geniepath}
Z.~Liu, C.~Chen, L.~Li, J.~Zhou, X.~Li, L.~Song, Y.~Qi, Geniepath: Graph neural
  networks with adaptive receptive paths, arXiv preprint arXiv:1802.00910.

\bibitem{zhuang2018dual}
C.~Zhuang, Q.~Ma, Dual graph convolutional networks for graph-based
  semi-supervised classification, in: Proceedings of the 2018 World Wide Web
  Conference on World Wide Web, International World Wide Web Conferences
  Steering Committee, 2018, pp. 499--508.

\bibitem{van2018filter}
D.~Van~Tran, N.~Navarin, A.~Sperduti, On filter size in graph convolutional
  networks, arXiv preprint arXiv:1811.10435.

\bibitem{xia2010multiview}
T.~Xia, D.~Tao, T.~Mei, Y.~Zhang, Multiview spectral embedding, IEEE
  Transactions on Systems, Man, and Cybernetics, Part B (Cybernetics) 40~(6)
  (2010) 1438--1446.

\bibitem{lin2012multi}
G.-F. Lin, H.~Zhu, C.-X. Fan, E.-H. Zhang, L.~Luo, Multi-cluster feature
  selection based on grassmann manifold, Jisuanji Gongcheng/ Computer
  Engineering 38~(16) (2012) 178--181.

\bibitem{turaga2011statistical}
P.~Turaga, A.~Veeraraghavan, A.~Srivastava, R.~Chellappa, Statistical
  computations on grassmann and stiefel manifolds for image and video-based
  recognition, IEEE Transactions on Pattern Analysis and Machine Intelligence
  33~(11) (2011) 2273--2286.

\bibitem{dong2013clustering}
X.~Dong, P.~Frossard, P.~Vandergheynst, N.~Nefedov, Clustering on multi-layer
  graphs via subspace analysis on grassmann manifolds, IEEE Transactions on
  signal processing 62~(4) (2013) 905--918.

\bibitem{wu2019comprehensive}
Z.~Wu, S.~Pan, F.~Chen, G.~Long, C.~Zhang, P.~S. Yu, A comprehensive survey on
  graph neural networks, arXiv preprint arXiv:1901.00596.

\bibitem{Veli2017Graph}
P.~Veličković, G.~Cucurull, A.~Casanova, A.~Romero, P.~Liò, Y.~Bengio, Graph
  attention networks, arXiv preprint arXiv:1710.10903.

\bibitem{2017arXiv171010568C}
J.~{Chen}, J.~{Zhu}, L.~{Song}, Stochastic training of graph convolutional
  networks with variance reduction, arXiv preprint arXiv:1710.10568.

\bibitem{Veli2018Deep}
P.~Veličković, W.~Fedus, W.~L. Hamilton, P.~Liò, Y.~Bengio, R.~D. Hjelm,
  Deep graph infomax, arXiv preprint arXiv:1809.10341.

\bibitem{2018arXiv180803965G}
H.~{Gao}, Z.~{Wang}, S.~{Ji}, {Large-Scale Learnable Graph Convolutional
  Networks}, arXiv preprint arXiv:1808.03965.

\end{thebibliography}

\end{document}